\documentclass[runningheads]{llncs}

\usepackage{eccv}

\usepackage{eccvabbrv}

\usepackage{graphicx}
\usepackage{booktabs}

\usepackage[accsupp]{axessibility}  % Improves PDF readability for those with disabilities.

\usepackage{graphicx}
\usepackage{amsmath}
\usepackage{amssymb}
\usepackage{booktabs}
\usepackage{multirow}
\usepackage{bm}
\usepackage{pifont}
\usepackage{subfloat}
\usepackage{marvosym}
\usepackage{hyperref}

\usepackage{orcidlink}

\newcommand{\authornote}[1]{{%
  \let\thempfn\relax% Remove footnote number printing mechanism
  \footnotetext[0]{#1}% Print footnote text
}}

\begin{document}

% ---------------------------------------------------------------
% TODO REVIEW: Replace with your title
\title{GTPT: Group-based Token Pruning Transformer for Efficient Human Pose Estimation} 

% TODO REVIEW: If the paper title is too long for the running head, you can set
% an abbreviated paper title here. If not, comment out.
\titlerunning{GTPT: Group-based Token Pruning Transformer}

% TODO FINAL: Replace with your author list. 
% Include the authors' OCRID for the camera-ready version, if at all possible.
\author{Haonan Wang\inst{1,2}$^*$\orcidlink{0000-0002-7159-2432} \and
Jie Liu\inst{1}\textsuperscript{\Letter}\orcidlink{0000-0002-9297-7729} \and
Jie Tang\inst{1}\orcidlink{0000-0002-6086-3559} \and
Gangshan Wu\inst{1}\orcidlink{0000-0003-1391-1762} \and
Bo Xu\inst{2}\orcidlink{0009-0006-2136-3814} \and
Yanbing Chou\inst{2}\orcidlink{0009-0006-1137-4771} \and
Yong Wang\inst{2}\orcidlink{0009-0001-2844-6296}}
% TODO FINAL: Replace with an abbreviated list of authors.
\authorrunning{H.~Wang et al.}
% First names are abbreviated in the running head.
% If there are more than two authors, 'et al.' is used.

% TODO FINAL: Replace with your institution list.
\institute{State Key Laboratory for Novel Software Technology, Nanjing University, China \and
Cainiao Network, China \\
\email{wanghaonan0522@gmail.com, \{liujie,tangjie,gswu\}@nju.edu.cn, \{songbai.xb,tonychou.zyb,Richard.wangy\}@cainiao.com}\\
% \url{http://www.springer.com/gp/computer-science/lncs} \and
% ABC Institute, Rupert-Karls-University Heidelberg, Heidelberg, Germany\\
% \email{\{abc,lncs\}@uni-heidelberg.de}
\textbf{\url{https://github.com/haonanwang0522/GTPT}}
}
% \institution{State Key Laboratory for Novel Software Technology,\\
%   Nanjing University,}
%   % \streetaddress{P.O. Box 1212}
%   \city{Nanjing}
%   % \state{Ohio}
%   \country{China}

\maketitle

\authornote{*: This work was conducted during Haonan Wang's internship at Cainiao Network.}
\authornote{\Letter: Corresponding author. liujie@nju.edu.cn}

\begin{abstract}
In recent years, 2D human pose estimation has made significant progress on public benchmarks. However, many of these approaches face challenges of less applicability in the industrial community due to the large number of parametric quantities and computational overhead. Efficient human pose estimation remains a hurdle, especially for whole-body pose estimation with numerous keypoints. While most current methods for efficient human pose estimation primarily rely on CNNs, we propose the Group-based Token Pruning  Transformer (GTPT) that fully harnesses the advantages of the Transformer. GTPT alleviates the computational burden by gradually introducing keypoints in a coarse-to-fine manner. It minimizes the computation overhead while ensuring high performance. Besides, GTPT groups keypoint tokens and prunes visual tokens to improve model performance while reducing redundancy. We propose the Multi-Head Group Attention (MHGA) between different groups to achieve global interaction with little computational overhead. We conducted experiments on COCO and COCO-WholeBody. Compared to other methods, the experimental results show that GTPT can achieve higher performance with less computation, especially in whole-body with numerous keypoints. 
\end{abstract}   
\section{Introduction}
\label{sec:intro}

2D human pose estimation (HPE)~\cite{chen20222d,wang2023lightweight,mao2022poseur,xu2022vitpose} is a core computer vision task that localizes human anatomical joints in images, underpinning applications such as 3D pose estimation~\cite{zeng2021learning,zou2021eventhpe,garau2021deca,wehrbein2021probabilistic}, activity recognition~\cite{vats2022key,yadav2022arfdnet}, and pose tracking~\cite{wang2020combining}. It is a field of significant interest in both academic and industrial sectors.

\begin{figure}[t]
		% \vspace{-0.2in}
		\begin{center}
			\includegraphics[width=0.45\linewidth]{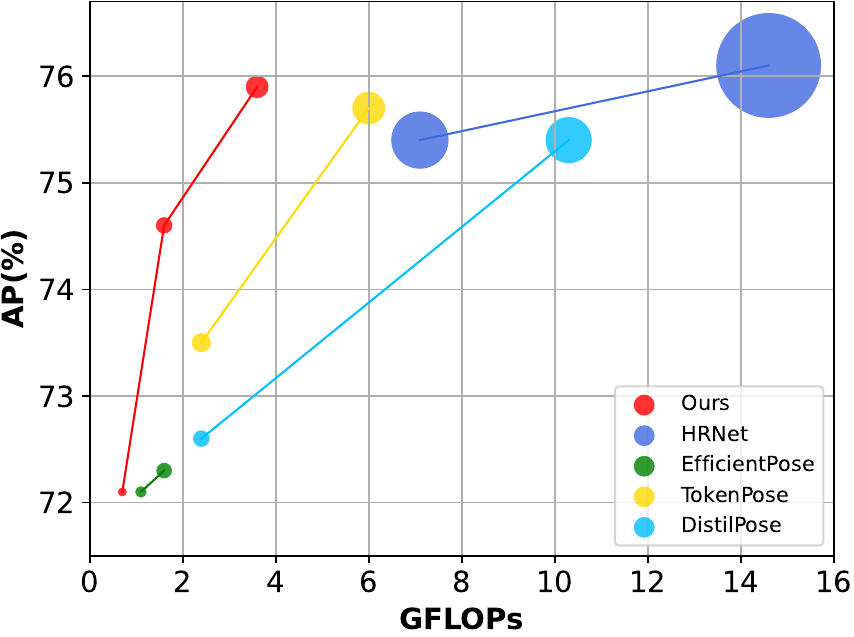}
		\end{center}
		\vspace{-0.2in}
		\caption{Comparison of our method with SOTA methods on COCO val. The horizontal coordinate indicates computation, the vertical coordinate indicates precision, and the circle size indicates the model parameters.}
		\label{fig:compare}
		\vspace{-0.3in}
	\end{figure}
 
    Despite achieving superior performance, 2D HPE still faces challenges in computation efficiency, particularly for whole-body pose estimation. Previous works on efficient pose estimation relied heavily on CNN-based methods to preserve the 2D structure. However, when the number of keypoints rises dramatically, high-resolution feature maps struggle to distinguish between dense keypoints, such as those on the face. Besides, keypoints are highly correlated and strongly constrained by dynamics and physics~\cite{tompson2014joint}. Unfortunately, CNN-based methods often struggle to capture the long-distance correlation between individual keypoints due to the limited receptive field.

    In recent years, Transformer has shown tremendous potential in various vision tasks, including pose estimation. TokenPose~\cite{li2021tokenpose} has effectively demonstrated the power of Transformer, as the attention mechanism can explicitly model relationships between keypoints. It conducts two tokenizations: keypoint tokens representing keypoint objects and visual tokens representing uniformly split image patches. Meanwhile, the Transformer can dynamically adjust the weights and accept variable-length inputs, resulting in fewer parameters and computational effort. Hence, we aim to leverage the superiority of the Transformer to develop an efficient pose estimation method. The previous method involves encoding all keypoints into keypoint tokens, which are concatenated with visual tokens and fed into the Transformer for feature extraction. 
    However, this approach introduces redundancy that negatively impacts efficiency. We investigate the redundancy of the model from two perspectives: keypoint tokens and visual tokens. 
    For keypoint tokens, redundancy increases significantly as the number of keypoints rises. It is particularly evident in body parts with a high density of keypoints. In shallow layers, keypoints within the same body part often focus on similar areas. Therefore, individually modeling each keypoint becomes highly redundant, especially when dealing with numerous keypoints.
    Besides, numerous visual tokens are redundant. In shallow layers, the attention of each keypoint token should primarily focus on the estimated human body rather than the background. As the model goes deeper, each keypoint token only focuses on a specific area around the keypoint, rather than the entire human body. 
    % Thus, we can prune the visual tokens that are inattentive.

\begin{figure}[t]
        % \vspace{-0.1in}
		\begin{center}
			\includegraphics[width=0.45\linewidth]{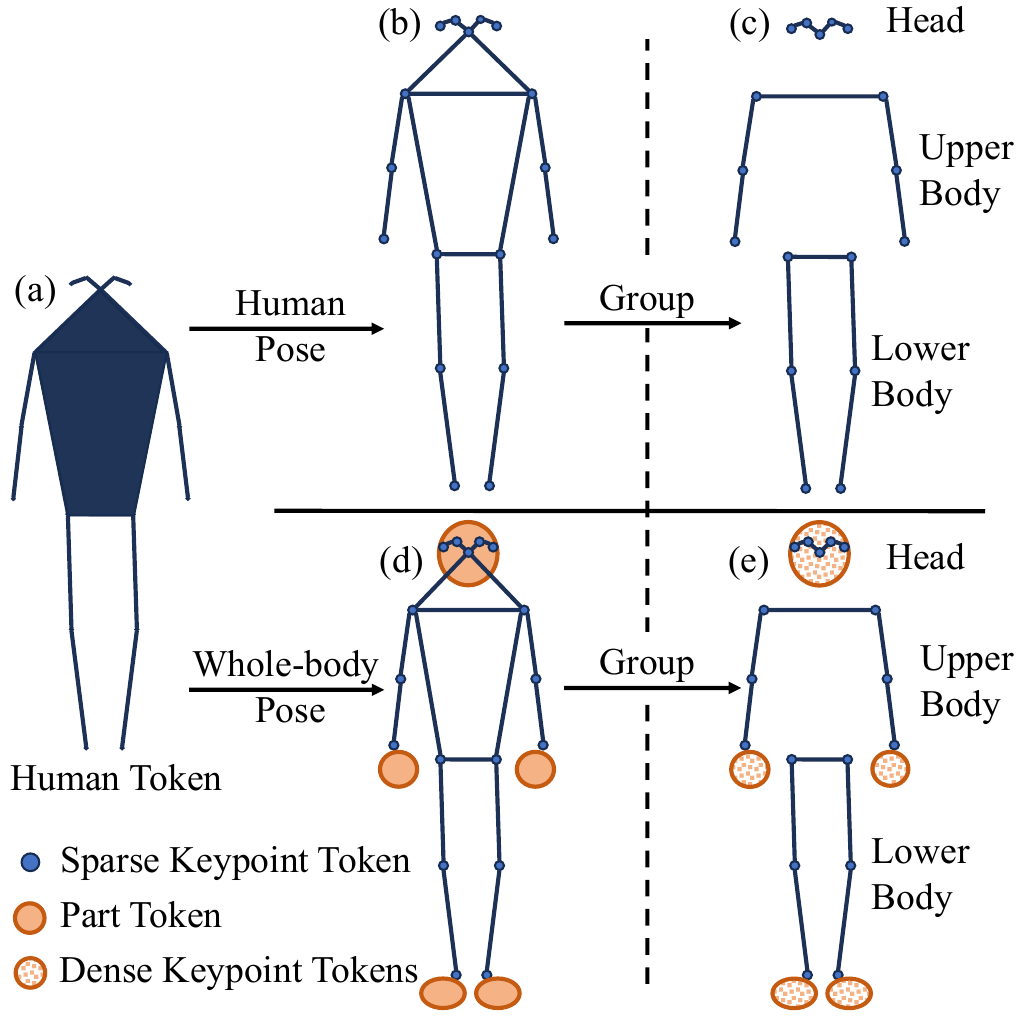}
		\end{center}
		\vspace{-0.2in}
		\caption{Overview of the introduction and grouping of keypoints. GTPT introduces keypoints in a coarse-to-fine manner. It starts with a human token and gradually transitions to sparse keypoint tokens and part tokens. Eventually, it converts part tokens into corresponding dense keypoint tokens. Besides, we categorize all keypoints into three groups: head, upper body, and lower body.}
		\label{fig:group}
		\vspace{-0.2in}
	\end{figure}

    Therefore, we propose a Group-based Token Pruning Transformer (GTPT) to enhance performance while minimizing redundancy. To mitigate the impact of numerous keypoints on efficiency without compromising performance, we propose a stepwise strategy to refine keypoints in a coarse-to-fine manner. Firstly, we categorize all the keypoints into two types: sparse keypoints on the body, and dense keypoints located elsewhere. As illustrated in \cref{fig:group} (a), only one human token is input initially. Subsequently, the human token gradually transforms into sparse keypoint tokens. For whole-body pose estimation, we introduce five part tokens (for the head, both hands, and both feet, as shown in \cref{fig:group} (d)) to be converted into dense keypoint tokens in the later stage. Besides, we employ pruning visual tokens to minimize redundancy. However, directly pruning with a high rate often prioritizes easily localizable keypoints while neglecting difficult-to-localize ones. To preserve visual tokens associated with keypoints, we propose group-based pruning, which enhances the model's performance and robustness to pruning. As depicted in \cref{fig:group} (c) and (e), we categorize keypoints into three groups: head, upper body, and lower body. To enhance performance, we model each group separately with masks. Furthermore, simply modeling keypoints within a group is insufficient for constraining the entire human body. Therefore, we propose the Multi-Head Group-Attention (MHGA), which not only models the keypoints within each group but also captures relationships among all keypoints while maintaining a low computational overhead.

Compared with previous works, our design offers three merits. (1) We boost performance with reduced computational overhead by group-based token pruning, as shown in \cref{fig:compare}. (2) GTPT introduces keypoints incrementally, following a coarse-to-fine strategy, thereby minimizing the impact of their increasing number on computation efficiency. (3) Our method intelligently prunes more visual tokens for areas with a higher density of keypoints, such as the face, while retaining more visual tokens for regions with fewer keypoints but more coverage, like the lower body, resulting in enhanced efficiency.  
 % (4) We employ self-distillation to optimize the non-differentiable pruning operation, effectively mitigating any adverse effects on performance.

The contributions are summarized as follows:
\begin{itemize}
    \item We propose a novel Group-based Token Pruning Transformer (GTPT) for efficient human pose estimation, especially whole-body pose estimation, which improves performance while reducing redundancy by grouping keypoint tokens and pruning visual tokens.
    \item We propose Multi-Head Group-Attention (MHGA) to model the keypoints within each group and capture the overall relationship between all the keypoints while keeping the computational overhead low.
    \item We propose a coarse-to-fine strategy to incrementally introduce keypoints, effectively mitigating the impact of numerous keypoints on efficiency.
\end{itemize}
\section{Related Work}
\label{sec:relatedWork}
% \subsection{Coordinate Classification}
%         Most previous methods for pose estimation represent keypoint locations in two ways. One approach is the regression-based method~\cite{li2021human,mao2022poseur,toshev2014deeppose} , which directly expresses keypoint positions as coordinates. The other approach is the heatmap-based method~\cite{xu2022vitpose,yuan2021hrformer,sun2019deep,huang2020devil,xiao2018simple,zhang2020distribution,wang2023lightweight}, which encodes positions as heatmaps for the model to learn from. While heatmap-based methods dominate in terms of performance in the pose estimation, they suffer from significant quantization errors. Thus, SimCC~\cite{li2022simcc} proposes to predict two 1D heatmaps instead of one 2D heatmap for each keypoint. Specifically, it treats the keypoint localization problem as a sub-pixel bin classification problem in horizontal and vertical coordinates. By maintaining a high resolution in the horizontal and vertical 1D heatmaps , the issue of quantization error is significantly alleviated. Besides, predicting two 1D heatmaps requires fewer parameters compared to a 2D heatmap for a keypoint token. Therefore, we incorporate SimCC into our lightweight pose estimation model.
 
\subsection{Transformer in HPE}
    
    Transformer~\cite{vaswani2017attention} has not only achieved remarkable success in Natural Language Processing (NLP) but has also shown strong performance in computer vision, including classification~\cite{dosovitskiy2020image,liu2021swin}, detection~\cite{zhu2020deformable}, and segmentation~\cite{cheng2021per}. Thus, pose estimation has also witnessed advancements, transitioning from CNNs to Transformers. We can broadly categorize previous works utilizing Transformers in HPE into two types: refining image features and aggregating part-wise features.

        \noindent
        \textbf{Refining image features.}  This manner enhances the feature extraction for images. For instance, TransPose~\cite{yang2021transpose} combines transformer encoder layers and CNN to capture global relationships effectively. HRFormer~\cite{yuan2021hrformer} replaces CNN with transformer encoders in HRNet~\cite{sun2019deep} to extract superior high-resolution features. TCFormer~\cite{zeng2022not} employs progressive clustering to merge tokens and preserve diverse scale information. ViTPose~\cite{xu2022vitpose} leverages a pre-trained SOTA transformer backbone to enhance accuracy.

        \noindent
        \textbf{Aggregating part-wise features.} Each keypoint is abstracted as a token to aggregate keypoint features. TFPose~\cite{mao2021tfpose} enhances CNN features with transformer encoders. Transformer decoders are then employed to extract keypoints features for regressing coordinates. Poseur~\cite{mao2022poseur} adopts deformable cross-attention to extract keypoints features. TokenPose~\cite{li2021tokenpose} feeds keypoint and visual tokens to the transformer for feature extraction, followed by predicting heatmaps using keypoint tokens. We incorporate transformers to query visual cues and learn anatomical constraints through attention. However, our approach differs from previous approaches in introducing keypoints in a coarse-to-fine manner. Furthermore, we strategically group keypoints to minimize computational costs while ensuring consistent performance, particularly in scenarios involving numerous keypoints, such as whole-body situations.

    \subsection{Efficient HPE}

    To improve efficiency, current work for efficient pose estimation concentrates on three key aspects: efficient architecture design, distillation, and pruning.
    
        \noindent
        \textbf{Efficient architecture design.} Many works focus on designing efficient architectures for this task~\cite{osokin2018real,neff2020efficienthrnet,wang2022lite,shen2021towards,bukschat2020efficientpose,yu2021lite}. EfficientPose~\cite{bukschat2020efficientpose} adopts neural architecture search (NAS) to obtain an efficient backbone network.	Besides, ZoomNAS~\cite{xu2022zoomnas} also employs a NAS framework to jointly search for connections between model architectures and different sub-modules, aiming to improve accuracy and efficiency. Lite-HRNet~\cite{yu2021lite} introduces shuffle blocks to HRNet and incorporates an efficient conditional channel weighting unit to replace the costly $1\times 1$ convolution in shuffle blocks. RTMPose~ \cite{jiang2023rtmpose} investigates factors influencing the performance and latency of pose estimation and develops a real-time model.
        
        \noindent
        \textbf{Distillation.} Knowledge distillation (KD)~\cite{hinton2015distilling} aims to transfer knowledge from teacher to student. FDP~\cite{zhang2019fast} firstly introduces KD into pose estimation. OKDHP~\cite{li2021online} proposes an online distillation approach that transfers pose structure knowledge in a one-stage manner. DistilPose~\cite{ye2023distilpose} introduces a heatmap-to-regression distillation framework that combines both benefits. DWPose~\cite{yang2023effective} proposes a two-stage distillation to enhance the performance of whole-body pose estimation.  
        % However, all previous works focus on migrating knowledge from a larger model to a smaller one to improve the latter's accuracy. In contrast, we focus on two models that share parameters, allowing us to optimize non-differentiable pruning operations.
        
        \noindent
        \textbf{Pruning.} The purpose of pruning is to reduce redundant visual tokens. PPT~\cite{ma2022ppt} proposes to localize the human body region and prune the background region by summing the attention maps of keypoint tokens to obtain a score. Besides, PPT adopts a pruning strategy that gradually increases the pruning ratio during training, which may negatively affect performance when the pruning ratio becomes too large. Therefore, we utilize a global perceived loss to transfer knowledge from the unpruned model to the pruned model.
        % To overcome the limitations of the pruned model's local information observation, we utilize a global-aware loss to transfer knowledge from the unpruned model, which has access to global information to enhance the pruned model's global awareness capability. 
        Besides, pruning may prune visual tokens associated with challenging keypoints, as it tends to focus on easily recognizable ones. Therefore, we introduce a grouping-based pruning method to retain visual tokens corresponding to all keypoints as much as possible.
         
%------------------------------------------------------------------------
% \clearpage

     \begin{figure}[ht]
		% \vspace{-0.1in}
		\begin{center}
			\includegraphics[width=0.8\linewidth]{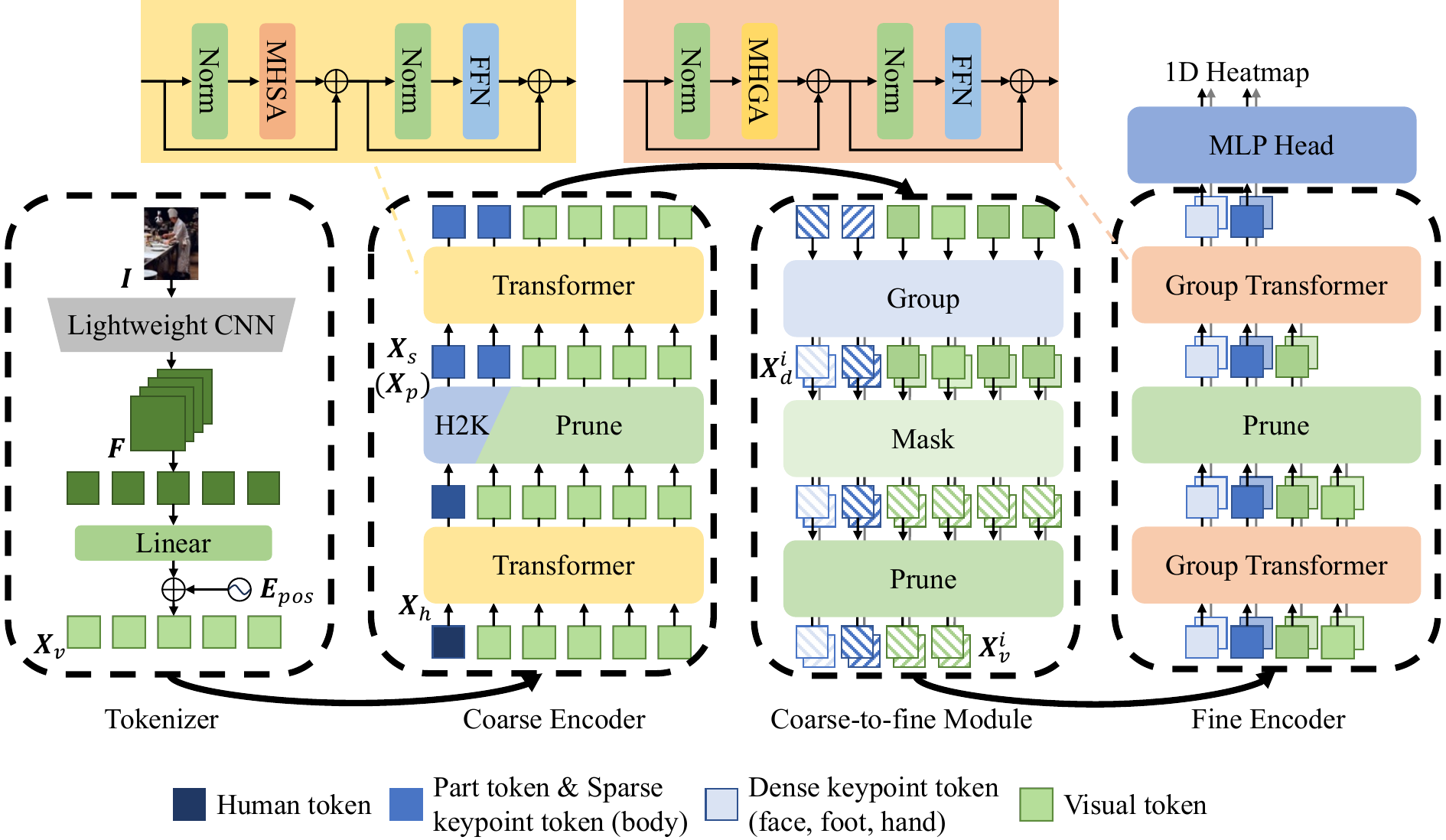}
		\end{center}
		\vspace{-0.2in}
		\caption{Overview of our proposed architecture. First, the tokenizer, which consists of a shallow CNN, performs feature extraction on the input image and transforms it into a sequence of tokens. Next, the Coarse Encoder, made of Transformer layers, gradually performs feature extraction from the target human to sparse keypoints as the network depth increases. The Coarse-to-Fine Module is the transition between the Coarse Encoder and the Fine Encoder, which introduces dense keypoints and groups the keypoints. To extract better features for each group, it masks the visual tokens differently in different groups. We then perform feature extraction on the keypoints of each group using the Fine Encoder. Finally, all keypoint tokens are fed into the unified MLP Head to estimate the 1D heatmaps of each keypoint.}
		\vspace{-0.2in}
    \label{fig:architecture}
	\end{figure}

\section{Method}

The key idea of our method is to take full advantage of the Transformer and improve computational efficiency by minimizing redundancy.
Thus, we propose a Group-based Token Pruning Transformer (GTPT). \cref{fig:architecture} illustrates the architecture, comprising the Tokenizer, Coarse Encoder, Coarse-to-fine Module, and Fine Encoder. GTPT effectively reduces redundancy by introducing keypoints in a coarse-to-fine manner and employing group-based pruning for visual tokens. 
\subsection{Tokenizer}

To meet the input requirements of Transformer~\cite{vaswani2017attention, dosovitskiy2020image}, a tokenizer is necessary to convert the 2D image $\bm{I}$ into a 1D sequence. Following TokenPose~\cite{li2021tokenpose}, the input image $\bm{I}$ is passed through a shallow CNN backbone to obtain the feature map $\bm{F}\in\mathbb{R}^{H\times W\times C}$, where $H$, $W$, and $C$ denote the height, width, and channel number. For more efficiency, different from TokenPose, we utilize ShuffleNet V2~\cite{ma2018shufflenet} as the CNN backbone. The feature map $\bm{F}$ is then uniformly divided into $N_{vis}=\frac{H}{P_H}\times\frac{W}{P_W}$ non-overlapping patches, where $(P_H, P_W)$ denotes the resolution of each patch. Each patch is flattened into a 1D vector of size $C\times P_H\times P_W$ and fed into a linear projection function to obtain the visual embedding $\bm{E}_v$. Since the Transformer is insensitive to position, but the human pose estimation task is sensitive to position, we incorporate position embeddings $\bm{E}_{pos}$ into visual embeddings $\bm{E}_v$ to obtain visual tokens $\bm{X}_v$ as follows:
        \begin{equation}
        \bm{X}_v = \bm{E}_v+\bm{E}_{pos}, \text{where } \bm{X}_v, \bm{E}_v, \bm{E}_{pos} \in\mathbb{R}^{N_{vis}\times d},
    \end{equation}
where $d$ is the dimension of hidden embeddings.

\subsection{Coarse Encoder}

The role of the coarse encoder is to extract shallow features for sparse keypoint tokens and part tokens to facilitate later grouping, pruning, and the introduction of dense keypoint tokens. It consists of Transformer layers, as shown in \cref{fig:architecture}.

 \noindent\textbf{Transformer Layer.}  The Transformer layer~\cite{vaswani2017attention} comprises two sub-layers: Multi-Head Self-Attention (MHSA) and Feed Forward Network (FFN). MHSA is a self-attention mechanism designed to capture dependencies between sequence elements. By attending to different aspects of sequence, MHSA effectively captures complex patterns and relationships between tokens. Specifically, for a sequence $\bm{X}$, it generates query $\bm{Q}$, key $\bm{K}$, and value $\bm{V}$ through three linear projections. Subsequently, we can perform attention operation:
\begin{align}
\label{eq:attn}
\text{Attention}\left(\bm{Q}_{i},\bm{K}_{i},\bm{V}_{i}\right) &= \bm{A}_i\bm{V}_{i} \\
&= \text{Softmax}\left(\frac{\bm{Q}_{i} \bm{K}_{i}^\mathrm{T}}{\sqrt{\hat{d}}}  \right)\bm{V}_{i}\;,
% \text{Attn}\left(\bm{Q}_{i},\bm{K}_{i},\bm{V}_{i}\right) = \text{SoftMax}\left(\frac{\bm{Q}_{i} \bm{K}_{i}^\mathrm{T}}{\sqrt{\hat{d}}}  \right)\bm{V}_{i}\;,
\end{align}
where $\bm{A}_i$ denotes the attention map of the $i^{th}$ head. The Multi-Head Attention performs attention operations simultaneously in $h$ different subspaces. The results are then concatenated and fed into a linear mapping to obtain outputs.

\noindent\textbf{Human-2-Keypoint.} In the shallow layers of TokenPose, each keypoint token focuses on the estimated human body, not the respective keypoints. To minimize computation, we only use a human token $\bm{X}_h\in\mathbb{R}^{1\times d}$ to extract the human features in the shallow layers. The input to the coarse encoder is $\bm{X}^0=[\bm{X}_h, \bm{X}_v]\in\mathbb{R}^{(N_{vis}+1)\times d}$, where $[\cdots]$ denotes the concatenate operation. In the middle of the coarse encoder, we convert the human token $\bm{X}_h$ into sparse keypoint tokens $\bm{X}_s$ and part tokens $\bm{X}_p$ through the Human-2-Keypoint (H2K): 
        \begin{equation}
        \bm{X}_s = \bm{X}_h+\bm{E}_s,\bm{X}_p = \bm{X}_h+\bm{E}_p,
    \end{equation}
where $\bm{E}_s,\bm{X}_s\in\mathbb{R}^{N_s\times d}$ denotes sparse keypoints' learnable embeddings and tokens, $\bm{E}_p,\bm{X}_p\in\mathbb{R}^{5\times d}$ denotes parts' learnable embeddings and tokens (for face, both hands, both feet) and $N_s$ denotes the number of sparse keypoints, which is the total number of body keypoints. Additionally, we employ a pruning technique to remove background visual tokens based on the attention maps of the human token to improve efficiency. Further details can be found in \cref{sec::prune}.

\subsection{Coarse-to-Fine Transition}

The Coarse-to-Fine Module plays a role in introducing dense keypoints, grouping all keypoints, and selecting appropriate masked visual tokens for each group. 

\noindent\textbf{Group Operation.} We group tokens to prune visual tokens more efficiently. In whole-body pose estimation, we incorporate the learnable dense keypoint embeddings $\bm{E}_d$ with corresponding part tokens to obtain dense keypoint tokens $\bm{X}_d$. Next, we divide all keypoints into three groups: head, upper body, and lower body, based on the location of each keypoint. Considering the definition of the whole-body pose estimation task, it is evident that the face has the most keypoints, but the smallest area. On the contrary, the lower body has the fewest keypoints, but the largest area. Consequently, after grouping, we can prioritize pruning more visual tokens for the face and retaining more for the lower body to ensure balanced computational efficiency for each group and enhance pruning efficiency without significantly compromising the performance.

\noindent\textbf{Mask Operation.} To ensure that visual tokens can adapt to the modeling requirements of each group, we enhance them through channel attention~\cite{hu2018squeeze}. Specifically, we employ an MLP to derive masks for each visual token as follows:
        \begin{equation}
        \bm{M}^0,\bm{M}^1,\bm{M}^2 = \text{Sigmoid}(\text{MLP}(\bm{X}_v)),
    \end{equation}
    where $\bm{M}^0, \bm{M}^1, \bm{M}^2\in\mathbb{R}^{N_{v}\times d}$ denotes the visual token masks corresponding to three groups, and $N_{v}$ denotes the number of remaining visual tokens. Afterward, we can obtain visual tokens for each group as follows:
        \begin{equation}
        \bm{X}_v^j = \bm{X}_v \cdot \bm{M}^j, 
    \end{equation}
where $j$ denotes the $j^{th}$ group, $\cdot$ denotes element-wise multiplication. 

\subsection{Fine Encoder}

The Fine Encoder consists of Group Transformer layers. We introduce a pruning layer in the middle of the Fine Encoder, reducing visual tokens in each group to enhance efficiency. Additionally, following SimCC\cite{li2022simcc}, we treat the keypoint localization problem as a classification task by discretizing the continuous horizontal and vertical axes into multiple bins. Consequently, we feed all keypoint tokens into the horizontal and vertical classifiers to predict 1D heatmaps.

\begin{figure}[t]
		% \vspace{-0.1in}
		\begin{center}
			\includegraphics[width=0.45\linewidth]{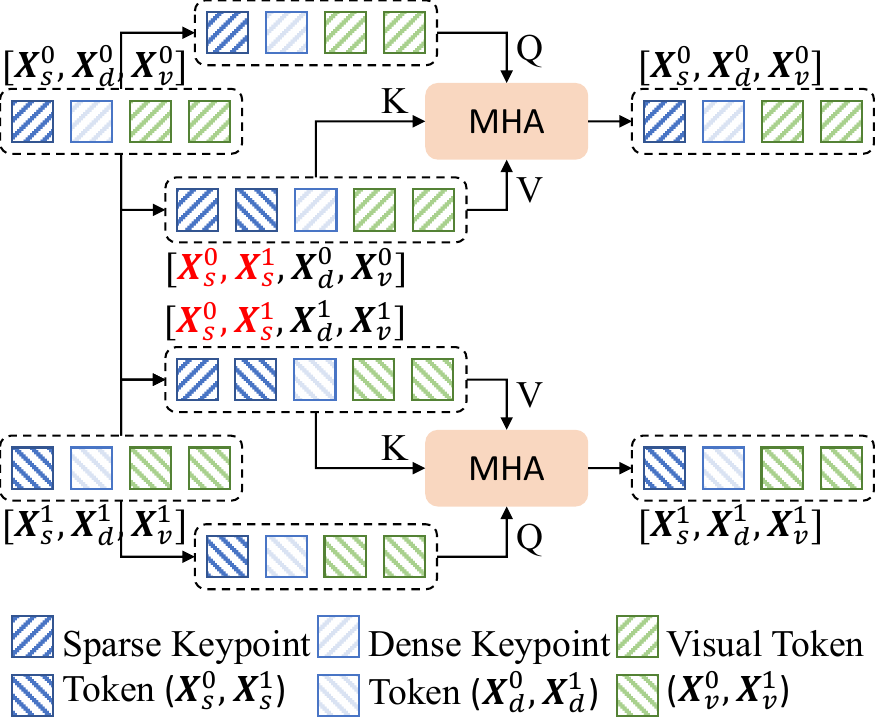}
		\end{center}
		\vspace{-0.2in}
		\caption{An example of the interaction between two groups via MHGA, where different patterns indicate different groups and red represents sharing.}
		\label{fig:mhga}
		\vspace{-0.2in}
	\end{figure}

\noindent
        \textbf{Group Transformer Layer.}  In the Group Transformer Layer, we introduce Multi-Head Group-Attention (MHGA) instead of Multi-Head Self-Attention. MHGA  aims to enhance computational efficiency by parallel computing attention only between the intra-group tokens. However, to enable interaction with inter-group information, we incorporate shared sparse keypoint tokens $\bm{X}_s$ as global information during the key and value calculation within each group. The specific process of MHGA is shown in \cref{fig:mhga}. For instance, when computing the MHGA of the $j^{th}$ group in the $i^{th}$ head, three linear projections are still employed to obtain the query $\bm{Q}_i^j$, key $\bm{K}_i^j$, and value $\bm{V}_i^j$: 
        % However, the inputs are no longer the same:
        \begin{align}
            &\bm{Q}_i^j=\bm{X}^j\bm{W}_i^Q\;,\\
            &\bm{K}_i^j=[\bm{X}_s,\bm{X}_d^j,\bm{X}_v^j]\bm{W}_i^K\;,\\ 
            &\bm{V}_i^j=[\bm{X}_s,\bm{X}_d^j,\bm{X}_v^j]\bm{W}_i^V\;. 
    \end{align}
The attention operation is then performed similarly as for the MHSA, utilizing  \cref{eq:attn}. The final output of the MHGA is obtained by concatenating and projecting the outcomes from the multiple heads.

\subsection{Global Perceived Pruning}
\noindent
\label{sec::prune}
    \begin{figure}[t]
    % \vspace{-0.15in}
        \begin{center}
            \includegraphics[width=0.45\linewidth]{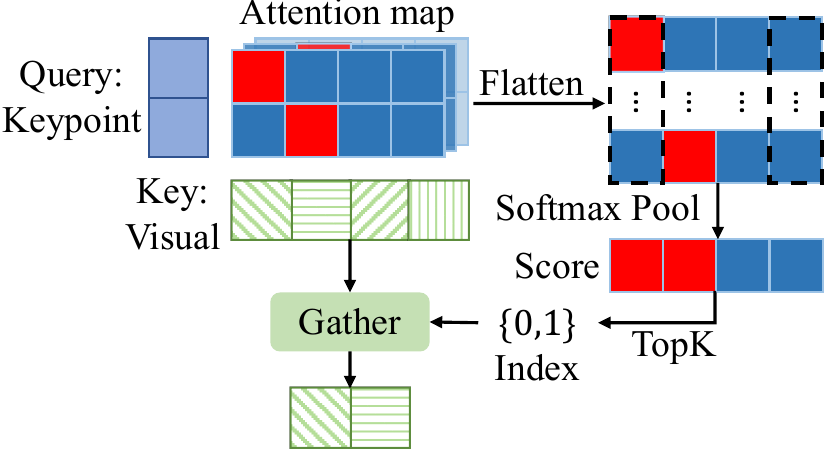}
        \end{center}
    \vspace{-0.2in}
    \caption{Overview of the pruning process, where different patterns represent different visual tokens. The red indicates high scores, and the blue indicates low scores.}
    \label{fig:prune}
    \vspace{-0.2in}
\end{figure}
\textbf{Pruning.} When visual tokens are input to estimate a person's pose, the attention is solely on that individual. Consequently, the visual tokens representing the background or other persons are redundant. 
% Furthermore, grouping leads to a notable replication of visual tokens. 
Therefore, it is crucial to prune redundant visual tokens to enhance efficiency. Inspired by PPT~\cite{ma2022ppt}, we utilize attention maps as the foundation for pruning. As shown in \cref{fig:prune}, attention maps employ the remaining visual tokens as keys during each pruning stage. However, each group takes their keypoint tokens as queries. Multiple attention maps exist due to the multi-head attention or multiple keypoint tokens. Therefore, pooling is necessary to calculate the importance score for each visual token. Since different heads compute attention scores in different subspaces, some significant visual tokens may only receive high attention scores in a few. Therefore, we propose to calculate the score through softmax pooling as follows:
\begin{align}
    \bm{S} = \sum_i^h\sum_k^{N_k}\bm{W}_i\bm{A}_i^k\;,\text{where } \bm{W}_i = \frac{e^{\bm{A}^{k}_i}}{\sum_i^h\sum_k^{N_k} e^{\bm{A}^{k}_i}}\;,
\end{align}
 $\bm{A}_i^k\in\mathbb{R}^{N_v}$ denotes the attention weight of the $k^{th}$ keypoint in the $i^{th}$ head, $\bm{S}\in\mathbb{R}^{N_v}$ denotes the importance scores, $h$ denotes the heads number, $N_v$ denotes the number of remaining visual tokens, and $N_k$ denotes the number of keypoints in the group. We can then select the top $\hat{N}_v$ visual tokens and prune the remaining ones. The number of retained visual tokens can be calculated as follows:
\begin{equation}
    \hat{N}_v=(1-\alpha)N_{vis}-N_d, 
\end{equation}
where $\alpha$ denotes the pruning rate,  $N_{vis}$ denotes the number of original visual tokens, and $N_{d}$ denotes the number of dense keypoint tokens within the group. Such a pruning technique ensures that the number of tokens in each group is approximately equal, thereby enhancing efficiency.

     \begin{figure}[h]
		% \vspace{-0.1in}
		\begin{center}
			\includegraphics[width=0.6\linewidth]{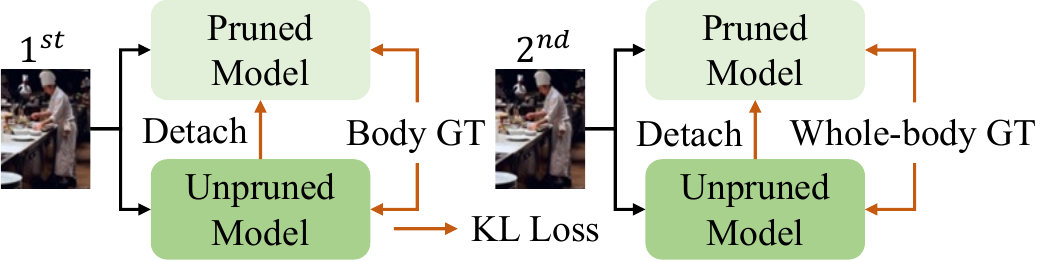}
		\end{center}
		\vspace{-0.2in}
		\caption{Overview of the training strategy. The left part indicates the first stage for HPE, and the right part indicates the second stage for whole-body pose estimation. }
		\label{fig:trainingstrategy}
		\vspace{-0.2in}
	\end{figure}
 
\noindent
        \textbf{Global Perceived Loss.}  
        % To mitigate the performance impact of high-rate pruning, we propose to utilize a globally aware loss. 
        Since the pruned model can only observe local information, it is easy to fall into local optimality. Therefore, we use pruned loss $L_{pruned}$ to supervise the pruned model while using unpruned loss $L_{unpruned}$ to supervise the unpruned model with shared parameters. Because the unpruned model can observe the global information, we transfer the knowledge from the unpruned model to the pruned model through the Global-2-Local loss $L_{G2L}$, enhancing the pruned model's global awareness. Thus, we train the model with the Global Perceived loss $L_{GP}$ as follows:
        % While the unpruned model can observe global information, the pruned model is limited to local information, yet shared parameters. During training (refer to \cref{fig:trainingstrategy}), we supervise both pruned and unpruned models by pruned loss $L_{pruned}$ and unpruned loss $L_{unpruned}$. Additionally, we transfer knowledge from the unpruned model to the pruned model through the Global-2-Local loss $L_{G2L}$, enhancing the pruned model's global awareness. Thus, we train the model with the globally aware loss $L_{GA}$ as follows:
        % we refer to the unpruned model as the teacher model and the pruned model as the student model. The only difference between the student and teacher models is whether they are pruned or not, and their parameters are shared. During training, as shown in \cref{fig:trainingstrategy}, we calculate the distillation loss $L_{distill}$ between the student's and teacher's logits, in addition to the respective losses $L_{tch}$ and $L_{stu}$ for the teacher and the student models. Thus, we can train the model with the total loss $L$ as:
        \begin{equation}
        L_{GP}=L_{pruned}+L_{unpruned}+L_{G2L}.
    \end{equation}

% \begin{table}[!t]%[!ht]
% \vspace{-0.1in}
% \centering
% \label{tab:COCOVal}
% \scalebox{0.85}{
% \input{data/coco_val}
% }
% \vspace{-0.1in}
% \caption{Comparison on COCO val. The input size is $256\times 192$.}
% \vspace{-0.25in}
% \end{table}
% \begin{table*}[!t]%[!ht]
% \vspace{-0.1in}
% \centering
% \label{tab:COCOWholeBodyVal}
% \scalebox{0.85}{
% \input{data/coco_wholebody_val}
% }
% \vspace{-0.1in}
% \caption{Comparison on COCO-WholeBody validation set.}
% \vspace{-0.2in}
% \end{table*}
% \begin{table}[!t]%[!ht]
% \centering
% \label{tab:MPII}
% \scalebox{0.85}{
% \input{data/mpii}
% }
% \vspace{-0.1in}
% \caption{Comparison on MPII validation. Input size is $256\times 256$.}
% \vspace{-0.3in}
% \end{table}
\noindent
        \textbf{Curriculum Learning.}  Since GTPT is a new architecture designed for efficient human pose estimation, it has not been pre-trained on large-scale datasets. Directly estimating the whole-body pose is a significant challenge due to the numerous keypoints involved. Therefore, we employ curriculum learning, enabling the model to gradually learn how to locate all keypoints of the whole body in an easy-to-difficult manner. As shown in \cref{fig:trainingstrategy}, we begin by training the model on simple body pose estimation. Then, we introduce part tokens and dense keypoint tokens to learn whole-body pose estimation. This approach effectively reduces the learning difficulty and enhances performance.

\begin{table}[!t]%[!ht]
% \vspace{-0.1in}
\centering
\scalebox{0.88}{
\begin{tabular}{lcccc}
\toprule
\textbf{Method} & \textbf{GFLOPs} & \textbf{Params} &\textbf{AP} & \textbf{AR}\\  \midrule
\multicolumn{5}{c}{\textbf{CNN-based Methods}}\\ \midrule
SimBa.-Res50~\cite{xiao2018simple} & 8.9   &34M   & 70.4 & 76.3 \\
SimBa.-Res101~\cite{xiao2018simple}  & 12.4 &53M & 71.4 & 77.1\\ 
SimBa.-Res152~\cite{xiao2018simple}  & 15.7 &68.6M & 72.0 & 77.8\\
HRNet-W32~\cite{sun2019deep}  & 7.1 &28.5M & 74.4 & 79.8\\  
HRNet-W48~\cite{sun2019deep}  & 14.6 &63.6M & 75.1 & 80.4\\  
Lite-HRNet-18~\cite{yu2021lite}  & 0.2 &1.1M & 64.8 & 71.2\\  
% Lite-HRNet-18~\cite{yu2021lite}  & 0.20 &1.1M & 64.8 & 71.2\\  
Lite-HRNet-30~\cite{yu2021lite}  & 0.3 &1.8M & 67.2 & 73.3\\  
% Lite-HRNet-30~\cite{yu2021lite}  & 0.31 &1.8M & 67.2 & 73.3\\  
EfficientPose-B~\cite{bukschat2020efficientpose}  & 1.1 &3.3M & 71.1 & -\\ 
EfficientPose-C~\cite{bukschat2020efficientpose}  & 1.6 &5.0M & 71.3 & -\\  \midrule

\multicolumn{5}{c}{\textbf{Transformer-based Methods}}\\ \midrule
TransPose-R-A4~\cite{yang2021transpose} & 8.9   &6.0M   & 72.6 & 78.0 \\
TransPose-H-S~\cite{yang2021transpose}  & 10.2 &8.0M & 74.2 & 79.5 \\
TokenPose-S-v1~\cite{li2021tokenpose} & 2.4   &6.6M   & 72.5 & 78.0 \\
% TokenPose-S-v1~\cite{li2021tokenpose} & 2.40   &6.6M   & 72.5 & 78.0 \\
TokenPose-B~\cite{li2021tokenpose} & 6.0   &13.5M   & 74.7 & 80.0 \\
% TokenPose-B~\cite{li2021tokenpose} & 6.00   &13.5M   & 74.7 & 80.0 \\
% TokenPose-L/D24~\cite{li2021tokenpose}  & 11.6 &27.5M & 75.8 & 80.9 \\
% TokenPose-L/D24~\cite{li2021tokenpose}  & 11.60 &27.5M & 75.8 & 80.9 \\
DistilPose-S~\cite{ye2023distilpose}  & 2.4   &5.4M   & 71.6 & - \\
% DistilPose-S~\cite{ye2023distilpose}  & 2.38   &5.4M   & 71.6 & - \\
DistilPose-L~\cite{ye2023distilpose}  & 10.3 &21.3M & 74.4 & - \\
% DistilPose-L~\cite{ye2023distilpose}  & 10.33 &21.3M & 74.4 & - \\
PPT-S~\cite{ma2022ppt}  & 2.0   &6.6M   & 72.2 & 77.8 \\
% PPT-S~\cite{ma2022ppt}  & 2.024   &6.6M   & 72.2 & 77.8 \\
PPT-B~\cite{ma2022ppt}  & 5.6   &13.5M & 74.4 & 79.6 \\\midrule
% PPT-B~\cite{ma2022ppt}  & 5.605   &13.5M & 74.4 & 79.6 \\\midrule

GTPT-T  & 0.7   & 2.4M   & 71.1 & 76.6 \\
GTPT-S  & 1.6   & 5.4M   & 73.6 & 78.9 \\
GTPT-B  & 3.6   & 8.3M   & 74.9 & 80.0 \\
 
 \bottomrule
\end{tabular}
}
% \vspace{-0.1in}
\caption{Comparison on COCO val. The input size is $256\times 192$.}
\vspace{-0.3in}
\label{tab:COCOVal}
\end{table}
\begin{table*}[!t]%[!ht]
% \vspace{-0.1in}
\centering
\scalebox{0.7}{
\begin{tabular}{lcc|cc|cc|cc|cc|cc}
\toprule
\multirow{2}{*}{\textbf{Method}} & \multirow{2}{*}{\textbf{Input Size}} & \multirow{2}{*}{\textbf{GFLOPs}} &\multicolumn{2}{c|}{\textbf{Whole-body}} & \multicolumn{2}{c|}{\textbf{Body}} & \multicolumn{2}{c|}{\textbf{Foot}} & \multicolumn{2}{c|}{\textbf{Face}} & \multicolumn{2}{c}{\textbf{Hand}}\\
\cmidrule(lr){4-5} \cmidrule(lr){6-7} \cmidrule(lr){8-9} \cmidrule(lr){10-11}  \cmidrule(lr){12-13}
& & & \textbf{AP} & \textbf{AR} & \textbf{AP} & \textbf{AR} & \textbf{AP} & \textbf{AR} & \textbf{AP} & \textbf{AR} & \textbf{AP} & \textbf{AR}\\\midrule
% \multicolumn{5}{c}{\textbf{CNN-based Methods}}\\ \midrule
SN~\cite{hidalgo2019single} & N/A   &272.3   & 32.7 & 45.6 & 42.7 & 58.3 & 9.9 & 36.9  & 64.9 & 69.7 & 40.8 & 58.0 \\
OpenPose~\cite{cao2021openpose}  & N/A &451.1 & 44.2 & 52.3 & 56.3 & 61.2 & 53.2 & 64.5 & 76.5 & 84.0 & 38.6 & 43.3 \\ 
PAF~\cite{cao2017realtime}  & $512\times 512$ & 329.1 & 29.5 & 40.5 & 38.1 & 52.6 & 5.3 & 27.8 & 65.6 & 70.1 & 35.9 & 52.8\\
AE~\cite{newell2017associative}  & $512\times 512$ & 212.4 & 44.0 & 54.5 & 58.0 & 66.1 & 57.7 & 72.5 & 58.8 & 65.4 & 48.1 & 57.4\\
DeepPose~\cite{toshev2014deeppose}  & $384\times 288$ & 17.3 & 33.5 & 48.4 & 44.4 & 56.8 & 36.8 & 53.7 & 49.3 & 66.3 & 23.5 & 41.0\\
SimBa.~\cite{xiao2018simple}  & $384\times 288$ & 20.4 & 57.3 & 67.1 & 66.6 & 74.7 & 63.5 & 76.3 & 73.2 & 81.2 & 53.7 & 64.7\\
HRNet~\cite{sun2019deep}  & $384\times 288$ & 16.0 & 58.6 & 67.4 & 70.1 & 77.3 & 58.6 & 69.2 & 72.7 & 78.3 & 51.6 & 60.4\\
PVT~\cite{wang2021pyramid}  & $384\times 288$ & 19.7 & 58.9 & 68.9 & 67.3 & 76.1 & 66.0 & 79.4 & 74.5 & 82.2 & 54.5 & 65.4\\
FastPose50-dcn-si~\cite{fang2022alphapose}  & $256\times 192$ & 6.1 & 59.2 & 66.5 & 70.6 & 75.6 & 70.2 & 77.5 & 77.5 & 82.5 & 45.7 & 53.9\\
ZoomNet~\cite{jin2020whole}  & $384\times 288$ & 28.5 & 63.0 & 74.2 & 74.5 & 81.0 & 60.9 & 70.8 & 88.0 & 92.4 & 57.9 & 73.4\\
ZoomNAS~\cite{xu2022zoomnas}  & $384\times 288$ & 18.0 & 65.4 & 74.4 & 74.0 & 80.7 & 61.7 & 71.8 & 88.9 & 93.0 & 62.5 & 74.0\\
ViTPose+-S~\cite{xu2022vitpose+}  & $256\times 192$ & 5.4 & 54.4 & - & 71.6 & - & 72.1 & - & 55.9 & - & 45.3 & -\\
ViTPose+-H~\cite{xu2022vitpose+}  & $256\times 192$ & 122.9 & 61.2 & - & 75.9 & - & 77.9 & - & 63.3 & - & 54.7 & -\\
RTMPose-m~\cite{jiang2023rtmpose}  & $256\times 192$ & 2.2 & 58.2 & 67.4 & 67.3 & 75.0 & 61.5 & 75.2 & 81.3 & 87.1 & 47.5 & 58.9\\
RTMPose-l~\cite{jiang2023rtmpose}  & $256\times 192$ & 4.5 & 61.1 & 70.0 & 69.5 & 76.9 & 65.8 & 78.5 & 83.3 & 88.7 & 51.9 & 62.8\\
\midrule

GTPT-T  & $256\times 192$ & 0.8  & 54.9 & 65.6 & 67.6 & 75.9 & 64.9 & 77.4 & 75.4 & 84.1 & 38.3 & 49.9 \\
GTPT-S  & $256\times 192$ & 2.0  & 59.6 & 69.9 & 71.0 & 78.7 & 70.4 & 82.2 & 81.0 & 87.6 & 45.4 & 57.0 \\
GTPT-B  & $256\times 192$ & 4.0  & 61.7 & 71.4 & 72.0 & 79.5 & 73.0 & 84.0 & 84.2 & 89.6 & 47.9 & 59.3 \\

 \bottomrule
\end{tabular}
}
% \vspace{-0.1in}
\caption{Comparison on COCO-WholeBody validation set.}
\vspace{-0.3in}
\label{tab:COCOWholeBodyVal}
\end{table*}
% \begin{table}[!t]%[!ht]
% \centering
% \scalebox{0.85}{
% \input{data/mpii}
% }
% % \vspace{-0.1in}
% \caption{Comparison on MPII validation. Input size is $256\times 256$.}
% % \vspace{-0.3in}
% \label{tab:MPII}
% \end{table}

\section{Experiments}
\subsection{Settings}
\label{exp:settings}

\noindent
\textbf{Dataset \& Evaluation Metrics.} COCO~\cite{lin2014microsoft} and COCO-WholeBody~\cite{jin2020whole} are the most popular for human and whole-body pose estimation, respectively. Both datasets contain the same images, but the labeling differs. They both contain more than $200,000$ images and $250,000$ human instances. Each person in COCO only labeled 17 keypoints, but COCO-WholeBody labeled more fine-grained keypoints on the face, hands, and feet, so labeled 133 keypoints for each person. We follow the standard splitting of \texttt{train2017} and \texttt{val2017}. There are $118k$ images in the train set and $5k$ in the validation set. Meanwhile, we utilize a two-stage top-down approach as the pipeline, which involves detecting the human by a detector first and then cropping it out to estimate its keypoints. In accordance with~\cite{xiao2018simple}, we employ a person detector with an accuracy of $56.4\%$ for the validation set. The standard evaluation metric for the COCO is the average precision (AP), which is calculated based on Object Keypoint Similarity (OKS). 
% Besides, to further validate the effectiveness of our proposed method, we conducted experiments on the MPII~\cite{lin2014microsoft}. The MPII dataset consists of approximately $25,000$ images,  each containing at least one person. In total, there are about $40,000$ humans in the dataset, each labeled with $16$ keypoints. For our experiments, we divided $28,000$ for training and the remaining for testing. To evaluate the performance, we used the head-normalized probability of correct keypoints (PCKh) metric.

% \noindent
% \textbf{Evaluation metric.} The standard evaluation metric for the COCO is the average precision (AP), which is calculated based on Object Keypoint Similarity (OKS):
% \begin{equation}
%     OKS=\frac{\sum_i\text{exp}(-d^2_{p^i}/2S^2_p\sigma^2_i)\delta(v_{p^i}>0)}{\sum_i\delta(v_{p^i}>0)},
% \end{equation}
% where $d_{p^i}$ represents the Euclidean distance between the $i^{th}$ keypoint of human $p$ and its corresponding ground truth, $v_{p^i}$ denotes the visibility of the keypoint, $S_p$ indicates the scale factor of human $p$, and $\sigma_i$ indicates the factor of keypoint $i$. 

\noindent
\textbf{Implementation Details.} For both training stages, we adopt the same setting. Specifically, we employ 4 GPUs to conduct experiments based on pytorch. Random initialization without pre-training necessitates extended training epochs for fitting GTPT better. We use Adam as the optimizer, which has a learning rate of $1\times10^{-3}$ decreased to $1\times10^{-4}$ at the $200^{th}$ epoch, to $1\times10^{-5}$ at the $260^{th}$ epoch and ended at the $300^{th}$; $\beta_1$ and $\beta_2$ are $0.9$ and $0.999$; weight decay is $10^{-4}$. 

\subsection{Results}
\label{exp:result}

To showcase the scalability of GTPT,we developed three model variants, ranging from the smallest to the largest: GTPT-T, GTPT-S, and GTPT-B.

\noindent
\textbf{COCO.} The results, presented in Table \cref{tab:COCOVal}, underscore the efficiency of our method. 
Traditional convolutional methods demand numerous parameters, high computational costs, and pre-training on large datasets for better performance. In contrast, GTPT-T and GTPT-B, despite similar performance to SimBa-Rse101 and HRNet-W48, require just $5\%$ and $25\%$ FLOPs, respectively, even with random initialization.
Although the efficiently designed Lite-HRNet and EfficientPose drastically reduce the number of parameters and computation. There is a significant loss in performance, and our method can achieve better results with a similar amount of computation. For example, the FLOPs of GTPT-S and EfficientPose-C are roughly the same, but we outperform them by $2.3$ AP. For the Transformer-based approach, it is possible to maintain a small number of parameters and still perform better in general. Therefore, our approach inherits this advantage and can get better results with fewer FLOPs. For example, when comparing GTPT-S with the pruned PPT-S and the unpruned TokenPose-S, our FLOPs are lower, but the performance is $1.1$ and $1.4$ AP higher.

\noindent
\textbf{COCO-WholeBody.} The results are shown in \cref{tab:COCOWholeBodyVal}. 
In the table, most methods are pre-trained on large-scale datasets (such as ZoomNAS and RTMPose), unlike GTPT, which uses random initialization. As a Transformer-based approach, GTPT’s complexity scales quadratically with input length. Therefore, we adopt $256 \times 192$ in GTPT for efficient whole-body estimation. 
% Despite this constraint, GTPT delivers impressive results.
% From the table, we can find the superiority of our method for whole-body pose estimation. 
Since group-based pruning
% , add them in a coarse-to-fine manner, and prune the redundant visual tokens
, we can achieve better results with less computational overhead. Compared to most methods that require high-resolution inputs, our approach achieves better results with $256\times 192$. GTPT-T achieves better results with only $15\% $ FLOPs to ViTPose+-S. Our method still has a comparative advantage over equally efficiently designed methods. For example, GTPT-S performs $1.4$ AP higher with lower FLOPs to RTMPose-m.

\noindent
% \textbf{MPII.} To validate the effectiveness of our proposed method, we performed experiments on the MPII dataset. The experimental results, summarized in \cref{tab:MPII}, illustrate the superiority of our method over other methods. For example, compared with SimpleBaseline-Res50, our method requires only 25\% parameters and 38\% computation to boost 2.0 PCKh. Furthermore, compared to other methods with comparable performance, the computational overhead of our method is lower. These findings strongly support the effectiveness of our proposed method in enhancing performance while maintaining computational efficiency.

\subsection{Ablation Study}
\label{exp:ablationStudy}

To exemplify the capability of our method in localizing a large number of keypoints, we uniformly adopt COCO-Wholebody as our dataset and employ group-based and global perceived pruning by default for ablation experiments.

\begin{table}[t]%[!ht]
\centering
% \vspace{-0.1in}
\scalebox{0.74}{
\begin{tabular}{c|ccccccc|c}
\toprule
 & \textbf{Group} & \textbf{Mask} & \textbf{GFLOPs} &\textbf{Body} & \textbf{Foot} & \textbf{Face} & \textbf{Hand} & \textbf{Whole}\\  \midrule

 \ding{172}  & \ding{55}  & \ding{55} & \textbf{1.77} & 69.2 & 68.0 & 79.8 & 44.3 & 58.2\\ 
 \ding{173}  & \checkmark & \ding{55} & 1.98 & 70.6 & 69.7 & 80.6 & 44.7 & 59.1(+0.9)\\
 \ding{174}  & \checkmark & \checkmark & 1.99 & \textbf{71.0} & \textbf{70.4} & \textbf{81.0} & \textbf{45.4} & \textbf{59.6(+1.4)}\\ 
 
 \bottomrule
\end{tabular}
}
% \vspace{-0.1in}
\caption{Ablation study of grouping and masking.}
\vspace{-0.3in}
\label{tab:group_mask}
\end{table}

\noindent
\textbf{Group \& Mask.}  The purpose of grouping and masking is to enhance the model's performance by extracting specific features for different groups. To validate their effectiveness, we conducted several ablation experiments. From the results in \cref{tab:group_mask}, we can conclude that excluding grouping (\ding{172} \vs \ding{173}) and masking (\ding{173} \vs \ding{174}) may reduce computation while leading to a drop in performance. However, when keypoints are grouped (\ding{172} \vs \ding{173}), the corresponding 0.9 AP performance improvement justifies the 0.21 increase in GFLOPs. Additionally, masking the visual tokens of different groups (\ding{173} \vs \ding{174}) results in a negligible increase in parameters and computation, yet it yields a remarkable 1.4 AP performance improvement compared to the baseline.

\begin{table}[t]%[!ht]
\centering
% \vspace{-0.1in}
\scalebox{0.8}{
\begin{tabular}{lccccc|c}
\toprule
\textbf{Method} & \textbf{GFLOPs} &\textbf{Body} & \textbf{Foot} & \textbf{Face} & \textbf{Hand} & \textbf{Whole}\\  \midrule

w/o Grouping & \textbf{1.77}  & 69.2 & 68.0 & 79.8 & 44.3 & 58.2\\ 
 MHSA & 1.99 & 70.2 & 69.4 & 80.4 & 44.0 & 58.8(+0.6)\\
 MHGA & 1.99 & \textbf{71.0} & \textbf{70.4} & \textbf{81.0} & \textbf{45.4} & \textbf{59.6(+1.4)}\\ 
 
 \bottomrule
\end{tabular}
}
% \vspace{-0.1in}
\caption{Ablation study of MHGA.}
\vspace{-0.25in}
\label{tab:mhga}
\end{table}

\noindent
\textbf{MHGA.} When we group keypoint tokens and adopt MHSA, the interaction between keypoints from different groups is limited, affecting the modeling of overall keypoint relationships. Therefore, we propose MHGA, which aims to interact with sparse keypoint tokens as global information while maintaining efficiency. To verify the effectiveness of MHGA, we conducted two experiments. The experimental results in \cref{tab:mhga} reveal that solely performing grouping with MHSA can only bring 0.6 AP improvement. However, MHGA improves by 1.4 AP compared to no grouping, highlighting the importance of information interaction among different groups. Meanwhile, the difference in computation efficiency between MHSA and MHGA is negligible. 

\noindent
\textbf{Keypoint Token Introduction Method.} To improve efficiency while maintaining performance, we propose a coarse-to-fine approach to introducing keypoint tokens, starting from 1 human token and gradually increasing to a small number of sparse keypoint tokens and part tokens, and then to a large number of dense keypoint tokens. To validate the effectiveness, we conducted three experiments. Upon reviewing \cref{tab:keypoint_token}, it is evident that the choice of the keypoint introduction approach has a negligible impact on the number of parameters. Directly introducing all keypoints yields the highest performance but incurs a higher computational overhead due to the excessive number of keypoints. In contrast, the Sparse-Dense approach effectively reduces computation with a slight impact on performance. Moreover, the Human-Sparse-Dense approach further decreases computation while rebounding in performance.

\begin{table}[t]%[!ht]
\centering
% \vspace{-0.1in}
\scalebox{0.65}{
\begin{tabular}{lcccccc|c}
\toprule
\textbf{Method} & \textbf{Params} & \textbf{GFLOPs} &\textbf{Body} & \textbf{Foot} & \textbf{Face} & \textbf{Hand} & \textbf{Whole}\\  \midrule

 Dense & \textbf{5.42M} & 2.26 & \textbf{71.0} & \textbf{70.7} & 80.8 & \textbf{45.7} & \textbf{59.8}\\ 
 Sparse-Dense & \textbf{5.42M} & 2.02(-11\%) & 70.8 & \textbf{70.7} & 80.7 & 45.0 & 59.5\\
 Human-Sparse-Dense & \textbf{5.42M} & \textbf{1.99(-12\%)} & \textbf{71.0} & 70.4 & \textbf{81.0} & 45.4 & 59.6\\ 
 
 \bottomrule
\end{tabular}
}
% \vspace{-0.1in}
\caption{Ablation study of keypoint token introduction method.}
\vspace{-0.3in}
\label{tab:keypoint_token}
\end{table}

% \begin{table}[!ht]%[!ht]
% \centering
% \vspace{-0.1in}
% \scalebox{0.75}{
% \input{data/curriculum}
% }
% \vspace{-0.1in}
% \caption{Ablation study of curriculum learning.}
% \vspace{-0.15in}
% \label{tab:curriculum_learning}
% \end{table}

% \noindent
% \textbf{Curriculum Learning.} To ease the challenge of learning whole-body pose estimation, we suggest implementing curriculum learning. This approach enables the model to gradually tackle increasingly complex tasks, starting with the localization of a few uniform keypoints before progressing to localizing numerous dense keypoints. To assess the necessity and effectiveness of curriculum learning in our model, we conduct ablation experiments. From \cref{tab:curriculum_learning},  the results highlight the significant improvement in accuracy for parts with fewer keypoints, such as body and feet, boosting their AP by 4.1 and 5.5. The reason is that if we directly learn the whole body keypoints, parts with more keypoints, such as face and hands, may dominate the optimization process, leading the model to fall into a local optimum. Consequently, the optimization of the parts with fewer keypoints is under-optimized.  

\begin{table}[t]%[!ht]
\centering
% \vspace{-0.1in}
\scalebox{0.72}{
% \begin{tabular}{lccc}
% \toprule
% \textbf{Pruning Optimization} & \textbf{GFLOPs} &\textbf{AP} & \textbf{AR}\\  \midrule

% Without Pruning & 3.05   & 73.8 & 79.0 \\
% Pruning Only  & 1.63(-47\%) & 73.3(-0.5) & 78.6(-0.4)\\ 
% Warm-up~\cite{ma2022ppt}  & 1.63(-47\%) & 72.6(-1.2) & 78.0(-1.0)\\
% Self-distill(ours)  & 1.63(-47\%) & 73.6(-0.2) & 78.9(-0.1)\\ 
 
%  \bottomrule
% \end{tabular}

\begin{tabular}{c|lcccccc|c}
\toprule
& \textbf{Pruning} & \textbf{G\&M} & \textbf{GFLOPs} &\textbf{Body} & \textbf{Foot} & \textbf{Face} & \textbf{Hand} & \textbf{Whole}\\  \midrule
% \ding{172}  & Direct  & \ding{55} & 1.77 & 67.4 & 65.5 & 78.5 & 41.9 & 55.9\\ 
\ding{172}  & Direct  & \ding{55} & 1.77 & 68.0 & 66.5 & 79.0 & 42.7 & 56.6\\ 
\ding{173}  & GP Loss  & \ding{55} & 1.77 & 69.2 & 68.0 & 79.8 & 44.3 & 58.2\\ \midrule
\ding{174} & Without & \checkmark & 3.53   & 70.9 & \textbf{70.7} & 80.8 & 44.9 & 59.4 \\
\ding{175} & Direct  & \checkmark & 1.99\textbf{(-44\%)} & 70.5 & 70.3 & 80.4 & 44.4 & 59.2\\ 
% Warm-up  & \textbf{1.99(-44\%)} & 69.7 & 69.3 & 80.3 & 43.6 & 58.3\\
\ding{176} & GP Loss  & \checkmark & 1.99\textbf{(-44\%)} & \textbf{71.0} & 70.4 & \textbf{81.0} & \textbf{45.4} & \textbf{59.6}\\ 
 
 \bottomrule
\end{tabular}
}
% \vspace{-0.1in}
\caption{Ablation study of group-based and global perceived pruning, where G\&M denotes grouping and masking.}
% , where warm-up denotes the method of optimizing pruning in the PPT~\cite{ma2022ppt}, i.e., starting with an unpruned model and gradually increasing the pruning ratio
\vspace{-0.25in}
\label{tab:pruning_opt}
\end{table}

\noindent
\textbf{Group-based and Global Perceived Pruning.} 
The pruned model is prone to fall into local optimality because it can only observe local information. However, the global awareness capability is essential for the pose estimation task, so we propose the global perceived loss. To verify the effectiveness of our proposed global perceived loss and whether grouping helps pruning, we conducted several experiments. \cref{tab:pruning_opt} reveals that pruning is crucial for efficiency (\ding{174} \vs \ding{176}), which directly reduces 44\% computation overhead. Notably, direct pruning with the proposed G\&M does not cause much performance drop (\ding{174} \vs \ding{175}). In contrast, direct pruning without G\&M results in significant performance loss (\ding{172} \vs \ding{173}), suggesting that our model exhibits a high tolerance for pruning. 
%Although direct pruning results in some performance loss (\ding{174} \vs \ding{175}), it is not substantial compared with without G\&M (\ding{172} \vs \ding{173}), suggesting that our model exhibits a high tolerance for pruning. 
When applying our proposed global perceived loss for pruning optimization, the model outperforms the unpruned model across most metrics (\ding{174} \vs \ding{176}), thus substantiating the effectiveness of global perceived loss.

To further validate that our group-based pruning eliminates redundant visual tokens, we visualize the attention maps in each layer before and after pruning.
Observation of the images reveals that the heatmap distributions are mostly unchanged after pruning, with the pruned visual tokens being confirmed as redundant, thus verifying the efficacy of our pruning approach. More detailed visualizations are provided in the supplementary materials.
\begin{figure}[t]
		% \vspace{-0.1in}
		\begin{center}
			\includegraphics[width=0.45\linewidth]{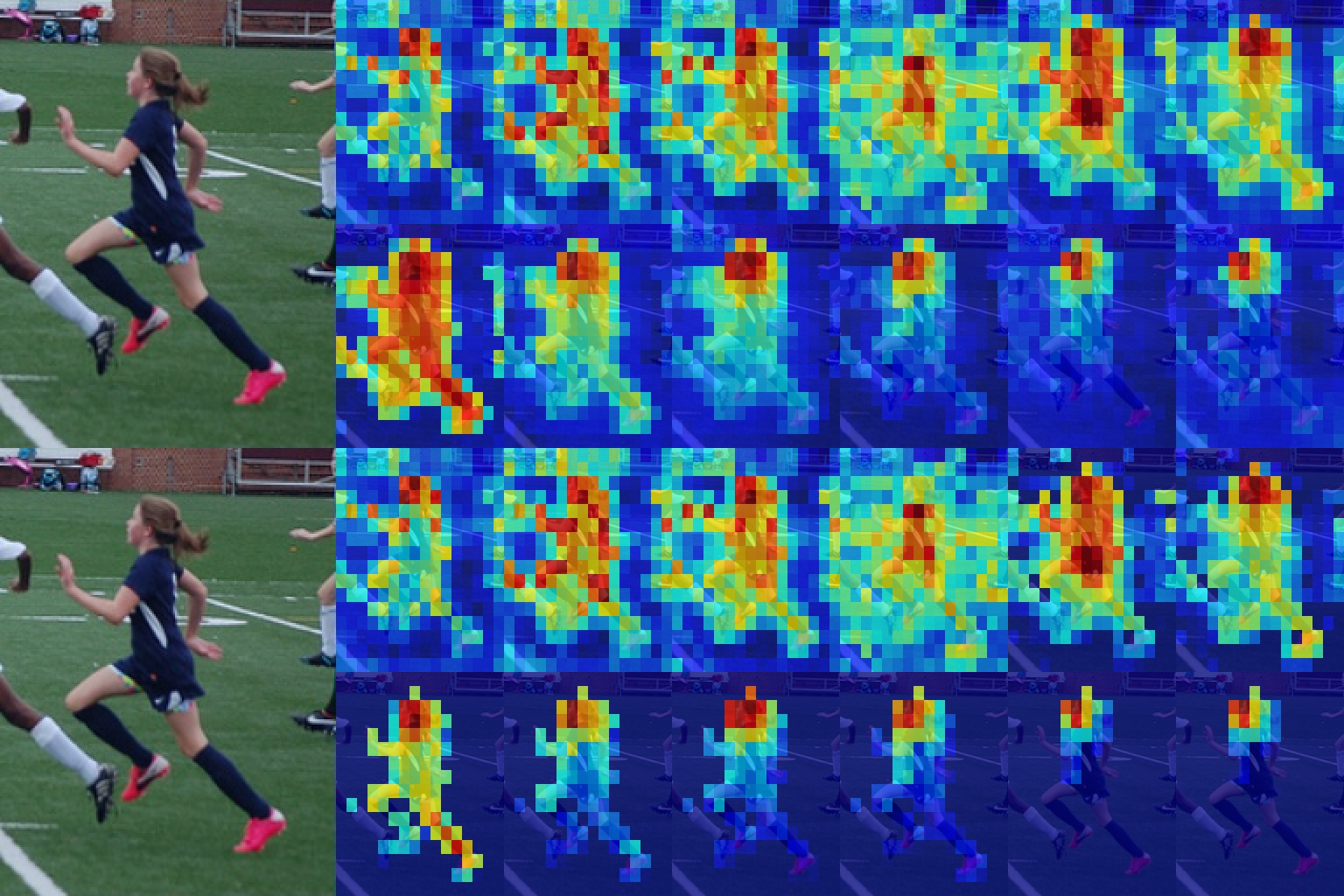}
		\end{center}
		\vspace{-0.2in}
		\caption{Attention map visualization for each layer.}
		\label{fig:mhga}
		\vspace{-0.25in}
	\end{figure}
% To verify whether the effectiveness of self-distillation can optimize pruning
% To determine the necessity and optimization requirements of pruning, we conducted various experiments, the results of which are presented in the \cref{tab:pruning_opt}. The table demonstrates the significance of pruning for human pose estimation. While the unpruned model exhibits FLOPs of 3.052, pruning reduces it to 1.626, showcasing a reduction of 47\%. Moreover, our model showcases a higher tolerance for pruning, with direct pruning resulting in a mere 0.5AP reduction.
% Furthermore, we proceed to compare the two optimization methods. The warm-up method suggested in the PPT does not apply to our model, as it leads to a decrease of 0.7AP. On the other hand, through self-distillation, our pruned model only experiences a marginal 0.2AP reduction compared to the unpruned model, making it a highly viable option.

\section{Limitations and Future Work}
\label{sec:limitations}

Though a nontrivial improvement has been achieved by GTPT, there are still some limitations worth exploring. First, GTPT, as a Transformer-based method, has a square-level relationship between its computational complexity and the length of the input sequence in attention computation. Therefore, high-resolution inputs lead to higher computation, more memory usage, and longer inference time. Thus, we choose $256 \times 192$ resolution as the input for efficient pose estimation. Second, CNNs have matured over time with extensive optimization, whereas Transformers, though rapidly advancing, are still evolving in algorithm optimization. As Transformer technology advances, new optimizations, like Flash-Attention~\cite{dao2022flashattention}, are expected to emerge, potentially speeding up our approach.

\section{Conclusion}
\label{sec:conclusion}

This work proposes the Group-based Token Pruning Transformer (GTPT) for efficient human pose estimation, especially whole-body pose estimation with numerous keypoints. GTPT reduces redundancy by grouping keypoint tokens and pruning visual tokens, improving performance while decreasing computation. To interact with inter-group information, GTPT performs global interaction using MHGA with less computational effort. To mitigate the impact of numerous keypoints, GTPT introduces keypoints in a coarse-to-fine manner, further reducing computational overhead. 
% Additionally, GTPT gradually introduces keypoints in a coarse-to-fine manner, further reducing computational overhead while maintaining performance. 
We conduct extensive experiments on COCO and COCO-WholeBody to demonstrate effectiveness.
\bibliographystyle{splncs04}
\bibliography{main}
\clearpage
\appendix

\clearpage
\setcounter{page}{1}

\section{Ablation study of Curriculum Learning}
\begin{table}[!ht]%[!ht]
\centering
% \vspace{-0.1in}
\scalebox{1.0}{
\begin{tabular}{cccccc|c}
\toprule
\textbf{CL} & \textbf{GFLOPs} &\textbf{Body} & \textbf{Foot} & \textbf{Face} & \textbf{Hand} & \textbf{Whole}\\  \midrule

 \ding{55} & \textbf{1.99} & 66.9 & 64.9 & 80.8 & 42.9 & 56.5\\
 \checkmark & \textbf{1.99} & \textbf{71.0} & \textbf{70.4} & \textbf{81.0} & \textbf{45.4} & \textbf{59.6(+3.1)}\\ 
 
 \bottomrule
\end{tabular}
}
% \vspace{-0.1in}
\caption{Ablation study of curriculum learning.}
% \vspace{-0.15in}
\label{tab:curriculum_learning}
\end{table}

\noindent To ease the challenge of learning whole-body pose estimation, we suggest implementing curriculum learning. This approach enables the model to gradually tackle increasingly complex tasks, starting with the localization of a few uniform keypoints before progressing to localizing numerous dense keypoints. To assess the necessity and effectiveness of curriculum learning in our model, we conduct ablation experiments. From \cref{tab:curriculum_learning},  the results highlight the significant improvement in accuracy for parts with fewer keypoints, such as body and feet, boosting their AP by 4.1 and 5.5. The reason is that if we directly learn the whole body keypoints, parts with more keypoints, such as face and hands, may dominate the optimization process, leading the model to fall into a local optimum. Consequently, the optimization of the parts with fewer keypoints is under-optimized. 

     \begin{figure}[htpb]
		\begin{center}
			\includegraphics[width=0.6\linewidth]{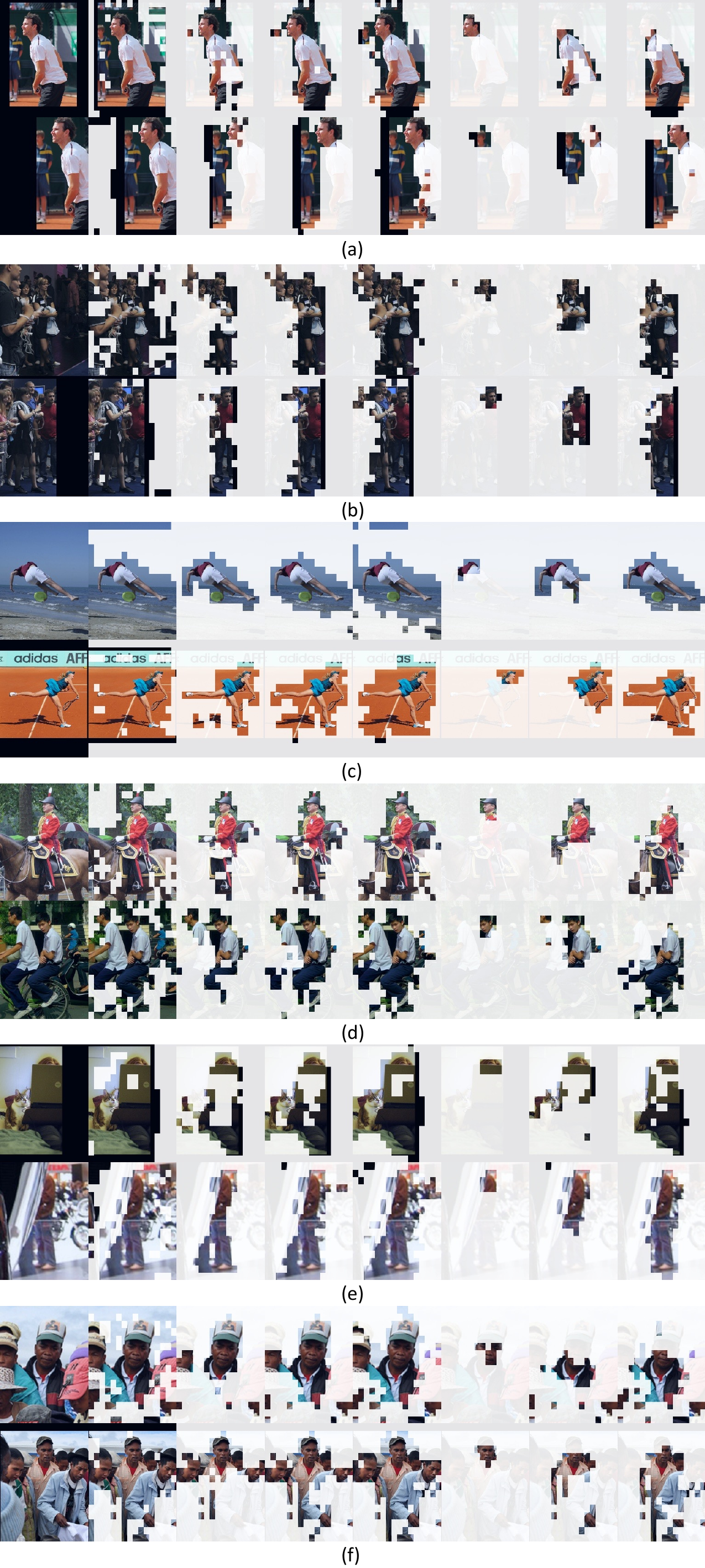}
		\end{center}
		% \vspace{-0.2in}
		\caption{Visualization of pruning results on COCO-WholeBody validation set. In each case, the first column shows the original image. The second column displays the result of the first pruning. The third to fifth columns represent the result of the second pruning, with the third, fourth, and fifth columns representing the three groups of the head, the upper body, and the lower body. The sixth to eighth columns represent the result of the third pruning, with the sixth, seventh, and eighth columns representing the three groups of the head, the upper body, and the lower body.}
		% \vspace{-0.25in}
    \label{fig:visPruning}
	\end{figure}
 
     \begin{figure}[htpb]
		\begin{center}
			\includegraphics[width=0.6\linewidth]{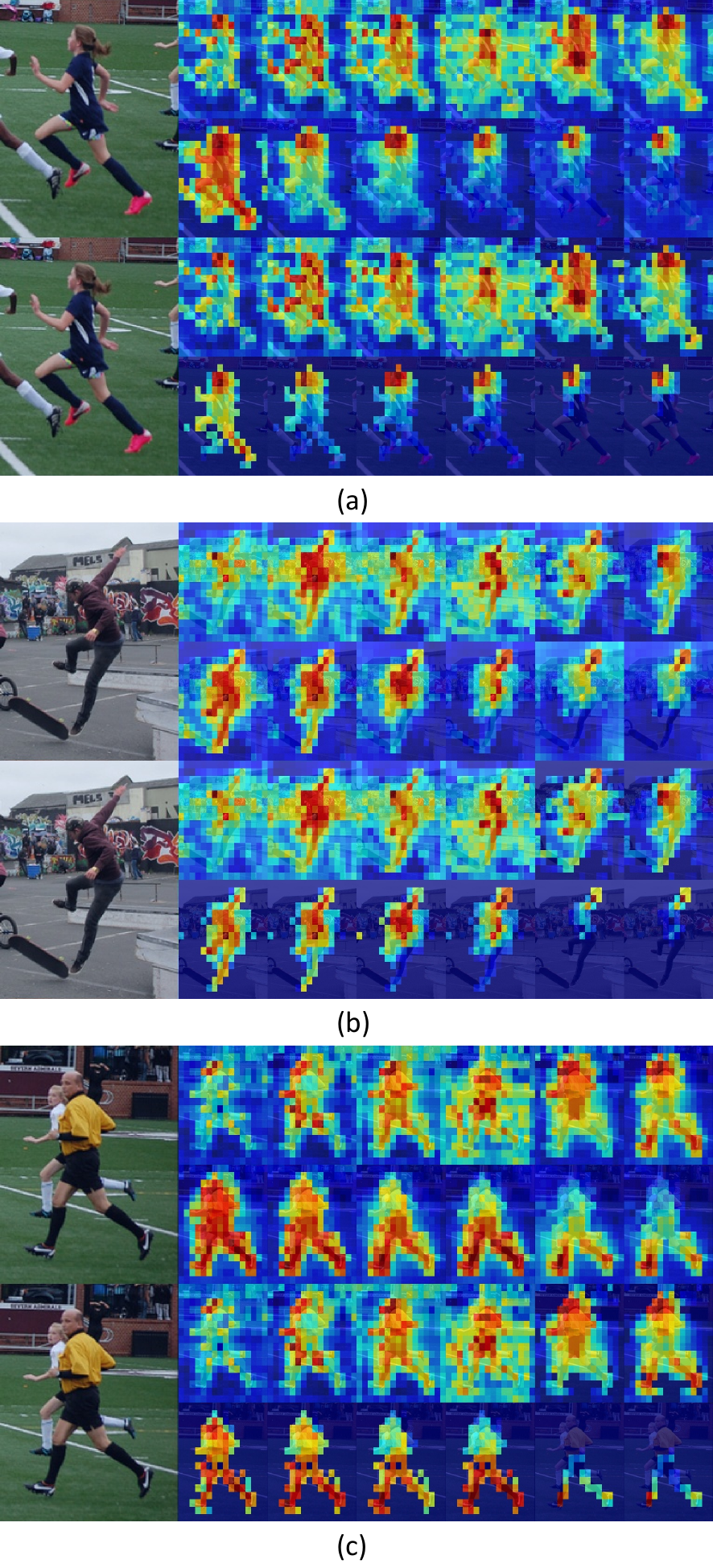}
		\end{center}
		% \vspace{-0.2in}
		\caption{Visualization of attention maps in different layers on COCO-Wholebody validation set. In each case, the $1^{st}$ row displays the attention maps for the $1^{st}$ through $6^{th}$ layers of the unpruned model. The $2^{nd}$ row displays the attention maps for the $7^{th}$ through $12^{th}$ layers of the unpruned model. The $3^{rd}$ row displays the attention maps for the $1^{st}$ through $6^{th}$ layers of the pruned model. The $4^{th}$ row displays the attention maps for the $7^{th}$ through $12^{th}$ layers of the pruned model. (a) shows the attention maps of the nose. (b) shows the attention maps of the wrist. (c) shows the attention maps of the ankle.}
		% \vspace{-0.25in}
    \label{fig:visAttentionDiffLayer}
	\end{figure}

\section{Visualizations}
\label{sec:visualizations}
\subsection{Pruning}
To confirm the redundancy of the pruned visual tokens, we visualize the pruning results by displaying the original images alongside the retained visual tokens for each group at each stage. \cref{fig:visPruning} demonstrates that the selected regions for each group become more refined as the network goes deeper. As anticipated, each group can focus on the visual tokens that correspond to its keypoints. Notably, even with a high pruning rate, the facial regions, which have numerous keypoints but smaller areas, are appropriately retained. And our method still works well in complex situations such as crowded scenes (\cref{fig:visPruning}(b)), different body poses(\cref{fig:visPruning}(c)), man-object interaction(\cref{fig:visPruning}(d)), occlusion(\cref{fig:visPruning}(e)), partial bodies(\cref{fig:visPruning}(f)), etc. In crowded scenes, as shown in \cref{fig:visPruning}(b), we can accurately prune out the currently estimated figure. When occluded, as shown in \cref{fig:visPruning}(e), our pruning preserves the unoccluded visual tokens within the group as much as possible. When there is only a partial body, as shown in \cref{fig:visPruning}(f), because of the grouping, our method can prune out the visual tokens of the corresponding region of the corresponding character contained within the current image.

\subsection{Attention Maps in different layers}
To further validate that our group-based pruning eliminates redundant visual tokens, we visualize the attention maps of selected keypoints before and after pruning in different layers. From the \cref{fig:visAttentionDiffLayer}, it is evident that during the initial stages of modeling, regardless of the keypoint, the attention is focused on the entire character's body rather than specific local keypoints. As the model deepens, the keypoints' attention gradually narrows down, focusing solely on the area surrounding each keypoint. Furthermore, we can observe that the pruned visual tokens are indeed redundant. The attention of keypoints focuses on mostly the same visual tokens before and after pruning. 

\begin{figure}[htpb]
    \begin{center}
        \includegraphics[width=0.6\linewidth]{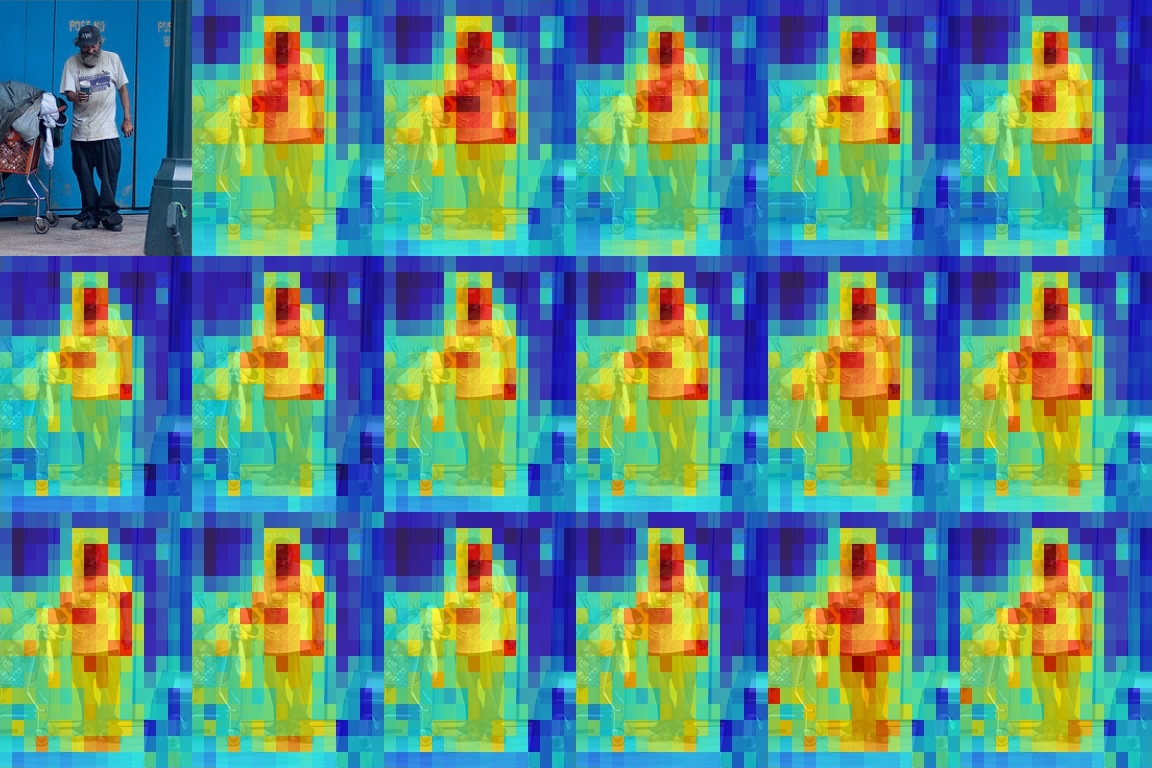}
    \end{center}
    % \vspace{-0.2in}
    \caption{Visualization of attention maps in the shallow($4^{th}$) layer on COCO-Wholebody validation set. Starting from the top left corner, 18 images are shown in order from left to right and top to bottom: original image, nose, left eye, right eye, left ear, right ear, left shoulder, right shoulder, left elbow, right elbow, left wrist, right wrist, left hip, right hip, left knee, right knee, left ankle, and right ankle.}
    % \vspace{-0.25in}
\label{fig:visAttentionShallow}
\end{figure}
 
\subsection{Attention Maps in the shallow layer}
To verify that the regions attended by different keypoints are similar in the shallow layer of the model, we visualized the attention maps of all body keypoints in the shallow layer. We uniformly chose the attention maps of the fourth layer for visualization. The visualization results are shown in Figure \ref{fig:visAttentionShallow}. As can be seen from the figure, in the shallow layer, all the keypoints focus on the human body, especially on the face. Therefore, the regions that different keypoints focus on are similar in the shallow layer.

\begin{table}[t]%[!ht]
\centering
\scalebox{0.8}{

\begin{tabular}{cccccc}
\toprule
\textbf{Model}& \textbf{Prune} & \textbf{GFLOPs} & \textbf{FPS} & \textbf{Throughput} & \textbf{Memory} \\  \midrule
GTPT-T& \ding{55}    & 1.33 & 93 & 928 & 1711M \\ 
GTPT-T& \checkmark & 0.83\textbf{(-46\%)} & 91 & 1775\textbf{(+91\%)} & 700M\textbf{(-59\%)}  \\ \midrule
GTPT-S& \ding{55}    & 3.53 & 93 & 701 & 1864M  \\ 
GTPT-S& \checkmark & 1.99\textbf{(-44\%)} & 91 & 1231\textbf{(+76\%)} & 784M\textbf{(-58\%)}  \\ \midrule
GTPT-B& \ding{55}    & 5.13 & 63 & 572 & 1886M  \\ 
GTPT-B& \checkmark & 4.02\textbf{(-22\%)} & 62 & 879\textbf{(+54\%)} & 838M\textbf{(-56\%)} \\
 \bottomrule
\end{tabular}
}
% \vspace{-0.1in}
\caption{Inference efficiency comparison on COCO-WholeBody validation. The input size is $256\times 192$.}
\vspace{-0.3in}
\label{tab:Efficiency}
\end{table}

\section{Inference Efficiency}

While GFLOPs already reflect the network's efficiency, it does not directly correspond to the actual runtime on the hardware. It is crucial to measure its actual runtime on the hardware to validate the efficiency of our model. The most commonly used metric for this purpose is FPS (Frames Per Second), which processes only one instance at a time. However, our approach follows a top-down framework. It means that by giving an input image, we first detect all instances of people through a detector, then crop, resize, and combine each human instance. Finally, we feed them into the pose estimation model using a mini-batch to accelerate inference. Therefore, we believe that evaluating the operational efficiency of the model in terms of throughput is more reasonable than using FPS. Throughput measures the maximum number of input instances that can be processed per unit of time, making it more consistent with the actual scenario when processing multiple instances in parallel. In addition, memory usage is also an indicator that we have to care about. On a graphics card with fixed memory, only the model with lower memory usage can handle more different human instances at the same time.

We conducted experiments by setting the batch size to 32, using pytorch, and performing inference on a single A100 GPU to control the variables. The results, including FPS, throughput, and memory usage, are presented in the \cref{tab:Efficiency}. Pruning does not reduce the inference time for a single instance. Because we need to sort the visual tokens based on their importance, select those with higher importance, and prune the ones with lower importance during the pruning process. The runtime saved by pruning is then utilized to compensate for the runtime consumed by the sorting process. However, pruning enhances the throughput and reduces its memory usage effectively. As a result, it improves the overall computational efficiency in practical applications. Besides, it is worth noting that our GTPT implementation utilizes PyTorch and does not incorporate efficiency-boosting technologies like FlashAttention~\cite{dao2022flashattention} that accelerate Transformer inference. It indicates that there is potential for further advancements in the efficiency of GTPT.

\end{document}